\documentclass{article}

 \usepackage[preprint]{neurips_2026}

\usepackage[utf8]{inputenc} 
\usepackage[T1]{fontenc}    
\usepackage{hyperref}       
\usepackage{url}            
\usepackage{booktabs}       
\usepackage{amsfonts}       
\usepackage{nicefrac}       
\usepackage{microtype}      
\usepackage{xcolor}         

\usepackage{algorithm}
\usepackage{algpseudocode}

\usepackage[utf8]{inputenc} 
\usepackage[T1]{fontenc}    
\usepackage{hyperref}       
\usepackage{url}            
\usepackage{booktabs}       
\usepackage{amsfonts}       
\usepackage{nicefrac}       
\usepackage{microtype}      
\usepackage{xcolor}         
\usepackage{comment}

\usepackage{multirow}
\usepackage{colortbl}
\usepackage{graphicx}
\usepackage{makecell}
\usepackage{utfsym}
\usepackage{graphicx} 
\usepackage{amsmath}
\usepackage{soul} 
\usepackage{caption}
\usepackage{adjustbox}
\usepackage{amssymb}
\usepackage{graphicx} 
\usepackage{bm} 
\usepackage{float}    
\usepackage{placeins}
\usepackage{natbib}
\setcitestyle{numbers,square}
\usepackage{pifont}
\usepackage{enumitem}

\usepackage{algorithm}
\usepackage{wrapfig}

\usepackage{appendix}

\definecolor{darkgreen}{RGB}{0,150,0}

\newcommand{\method}{IRR-Drive}

\definecolor{myblue}{RGB}{47,85,151}
\definecolor{myred}{RGB}{242,160,120}
\newcommand{\graycolorrow}[1]{\rowcolor{gray!15} #1}
\newcommand{\suppltext}[1]{\textbf{\textcolor{blue}{Supplementary Material #1}}}

\definecolor{lightblue}{RGB}{240,248,255}
\newcommand{\colorrow}[1]{\rowcolor{lightblue} #1}

\title{Intend, Reflect, Refine: An Adaptive Multimodal Reflection Framework for Autonomous Driving}

\author{%
    \textbf{
      Zisheng Chen$^{1,}$\thanks{These authors contributes equally},
      Yuping Qiu$^{2, \ast}$,
      Jianhua Han$^{3}$,
      Tao Tang$^{1}$,
      Xiuwei Chen$^{1}$,
      }\\
    \textbf{
      Likui Zhang$^{1}$,
      Ying-Cong Chen$^{2}$,
      Hang Xu$^{3}$,
      Xiaodan Liang$^{1,}$\thanks{Corresponding Author}
    }\\[0.3cm]
    $^1$ Sun Yat-sen University \quad
    $^2$ HKUST(GZ) \\
    $^3$ Yinwang Intelligent Technology Co. Ltd. \\
}

\begin{document}

\maketitle

\begin{center}
    \includegraphics[width=0.98\textwidth]{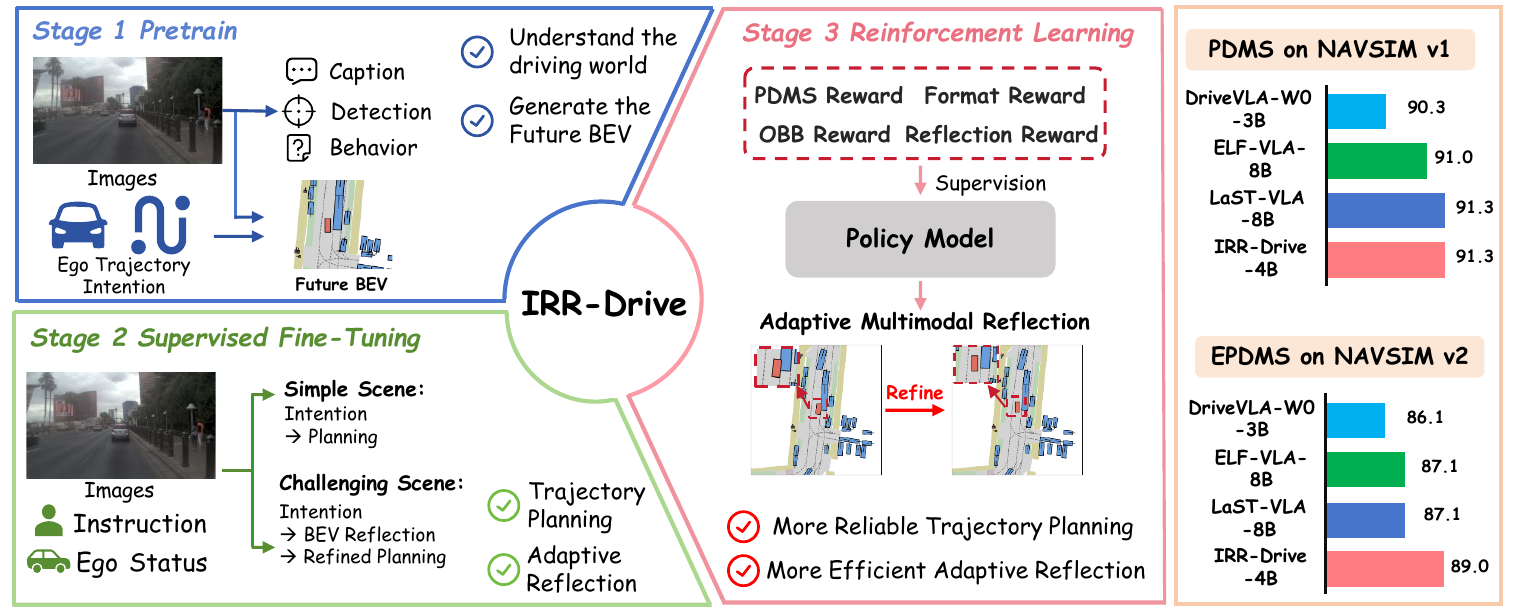}
    \captionof{figure}{
    Visual overview of IRR-Drive. 
    The model learns through three progressive stages. 
    We first pretrain the model on driving world understanding and future BEV generation to establish foundational perception and future forecasting capability. 
    Then, we fine-tune the model, applying direct execution for simple scenes and BEV-based reflection for challenging ones to learn adaptive reasoning. 
    Finally, we use reinforcement learning, incorporating multiple reward functions to refine the policy, enabling more reliable trajectory planning and more efficient adaptive reflection.
    The results on the right show that IRR-Drive-4B achieves highly competitive PDMS and superior EPDMS scores on NAVSIM benchmarks despite a smaller parameter scale.
    }
    \label{fig:teaser}
\end{center}

\begin{abstract}
\label{sec:abs}
Recent Vision-Language-Action (VLA) models have advanced end-to-end autonomous driving by incorporating reasoning for better interpretability and planning quality. 
However, most existing approaches directly generate the final trajectory without explicitly examining its future consequences, which limits their reliability in complex and dynamic environments.
To address this limitation, we propose \textbf{\method{}} (\textbf{Intend, Reflect, Refine}), an adaptive multimodal reflection framework for autonomous driving. 
Specifically, to tightly couple high-level reasoning with physical constraints, \method{} first generates an initial textual trajectory intention and anticipates potential interactions by predicting future semantic bird's-eye view (BEV) representations. 
This dual-modal (Text + BEV) reflection space explicitly models anticipated scene evolution, enabling the model to perform grounded self-correction and refine its initial intent before generating the final trajectory. 
Furthermore, to balance planning performance and computational efficiency, we construct reflection-oriented training data and design an adaptive reflection reward, enabling the model to adaptively select its reasoning mode according to scene complexity.
Instead of using reasoning primarily as an auxiliary interpretation, \method{} directly integrates an adaptive reflection mechanism into the planning framework, enabling grounded, decision-aware trajectory correction that is driven by scene complexity.
Our method achieves state-of-the-art performance on the NAVSIM benchmark in both PDMS and EPDMS.
Extensive experiments demonstrate the effectiveness of our multimodal reflection framework and validate the efficacy of the proposed adaptive reflection strategy.

\end{abstract}

\section{Introduction}

Recent Vision-Language-Action (VLA) approaches have advanced end-to-end autonomous driving by integrating perception, reasoning, and planning within a unified framework~\cite{AdaThinkDrive, AutoVLA, autodrivepi, fsdrive}. 
By incorporating reasoning mechanisms, these models improve interpretability and enable more informed decision-making than purely reactive policies.
However, the majority of current methods produce trajectories in a single pass, lacking an explicit assessment of how candidate actions will shape future states.
Effective driving requires not only producing a feasible trajectory, but also evaluating and revising it before execution.
This motivates reflective planning, where candidate actions are assessed and refined according to their anticipated outcomes.
Despite recent progress, current reasoning paradigms remain limited in supporting such grounded reflective decision-making.

A key challenge lies in the lack of a suitable representation space for plan verification and correction.
Existing reasoning-based driving methods~\cite{elfvla, AdaThinkDrive, AutoVLA} often express intermediate reasoning through textual or symbolic abstractions, which are interpretable but too coarse to capture fine-grained spatial interactions required for trajectory refinement.
In contrast, vision-based approaches~\cite{fsdrive,last-vla} preserve richer physical details, yet typically use them as implicit features or auxiliary supervision rather than as an explicit space for evaluating planned actions.
As a result, reasoning remains weakly grounded in the physical consequences of actions, limiting its effectiveness in complex driving scenarios.

To address these limitations, we propose \textbf{\method{}} (\textbf{Intend, Reflect, Refine}), an action-conditioned multimodal framework with adaptive reflection for autonomous driving.
Given an input scene, \method{} first generates an initial trajectory intention.
Conditioned on both the current scene and this intention, it predicts future semantic bird's-eye view (BEV) representations, which serve as a structured reflection space for analyzing future interactions and risks.
The model then refines the initial intention based on this multimodal reflection and produces the final trajectory.
To improve efficiency, \method{} further learns adaptive reflection behavior, directly outputting trajectories in simple scenes while invoking reflective refinement in challenging ones.
Thus, reflection is no longer a static post-hoc explanation, but an adaptive, decision-aware process for trajectory correction.

In implementation, we first pretrain the model on driving understanding data and semantic BEV generation data, then construct NAVSIM-derived~\cite{navsimv1} non-reflection and multimodal reflection data for supervised fine-tuning.
This enables the model to operate in both non-reflective and reflective reasoning modes.
Finally, we apply GSPO reinforcement learning with rewards that jointly consider planning quality, trajectory alignment, and reflection efficiency, encouraging a better trade-off between closed-loop performance and computational cost.
Our main contributions are summarized as follows:
\begin{itemize}[nosep, leftmargin=12
pt]
    \item  We design a multimodal reflection module that integrates textual reasoning with semantic BEV prediction, providing a grounded space for trajectory verification and refinement.
    \item We develop an adaptive reflection strategy that selects direct prediction or reflective refinement based on scene complexity, improving efficiency while preserving planning quality.
    \item We conduct extensive experiments on NAVSIM v1 and v2, demonstrating strong closed-loop planning performance and validating the effectiveness of multimodal reflection, adaptive mode selection, and reinforcement learning.
\end{itemize}

\section{Related Work}

\textbf{Reasoning and Self-Reflection in VLA.}
Building on the VLA paradigm, recent work has placed increasing emphasis on explicit reasoning in driving decision-making. FSDrive ~\cite{fsdrive} replaces purely textual reasoning with visual spatio-temporal chain-of-thought by generating future visual states, thereby preserving finer spatial details during deliberation. 
AutoVLA ~\cite{AutoVLA} unifies reasoning and action generation within a single autoregressive framework and incorporates adaptive reasoning together with reinforcement fine-tuning to improve planning performance. 
AdaThinkDrive ~\cite{AdaThinkDrive} introduces fast and slow reasoning modes, allowing the model to allocate inference depth according to scene complexity. 
Beyond descriptive reasoning, self-reflective planning has also emerged as an important direction. 
Counterfactual VLA (CF-VLA) ~\cite{cvla} combines time-segmented meta-actions with counterfactual analysis to revise planned behaviors based on simulated outcomes. 
ELF-VLA ~\cite{elfvla} leverages teacher-generated structured diagnoses and corrected trajectories to perform failure-driven refinement during training.

Additional related works on \textit{Vision-Language and Vision-Language-Action Models for Autonomous Driving} and \textit{Future Scene Modeling for Autonomous Driving} are provided in the \suppltext{~\ref{app:related_work}}. 

\section{Method}

In this section, we describe \textbf{\method{}}.
Sec.~\ref{subsec-tokenizer} presents the semantic BEV tokenizer, which converts BEV images into discrete codes.
Sec.~\ref{subsec-data} describes the construction of pretraining, supervised fine-tuning, and RL training data.
Sec.~\ref{subsec-sft} introduces the two-stage supervised fine-tuning procedure, and Sec.~\ref{subsec-rl} presents reinforcement learning for improving planning performance and adaptive reflection.

\subsection{Semantic BEV Tokenizer}
\label{subsec-tokenizer}
Inspired by autoregressive image generation paradigms~\cite{vqgan,vqvae} and unified vision-language models (VLMs)~\cite{illume+, semhitok, liquid}, we develop a semantic BEV tokenizer that discretizes BEV into code tokens, thereby enabling unified autoregressive reasoning and planning.

Specifically, given a BEV representation ${B} \in \mathbb{R}^{H \times W \times C}$, we first extract semantic features using a frozen encoder $E_{\theta}(\cdot)$: ${F} = E_{\theta}({B})$.
The features are projected via an MLP $g_{\phi}(\cdot)$ into a quantization space: ${H} = g_{\phi}({F})$.
We then apply vector quantization with a learnable codebook ${Z} = \{ {z}_k \}_{k=1}^K$, where feature is mapped to its nearest code:
\begin{equation}
\mathbf{Z_q}, {I_q} = \mathop{\arg\min}\limits_{k \in \{1,\dots,K\}} \| {H} - \mathbf{Z[k]} \|_2^2
\end{equation}
Where $K$ is the codebook size, $\mathbf{Z_q}$ is the quantized feature, and ${I_q}$ is the quantized index.
The quantized features are then decoded by $D_{\psi}(\cdot)$ to reconstruct the semantic representation: $\hat{F} = D_{\psi}({Z}_q)$.
The tokenizer is trained with a standard VQ objective:
\begin{equation}
    \mathcal{L} = \mathcal{L}_{sem} (F,\hat{F}) + \| \text{sg}[H] - Z_q \|_2^2 + \beta \| \mathbf{H} - \text{sg}[\mathbf{Z}_q] \|_2^2
\end{equation}
where $\mathcal{L}_{sem}$ represents the mean squared error loss, $\text{sg}[\cdot]$ denotes the stop-gradient operator.

The semantic BEV tokenizer converts BEV representations into discrete codes, thereby enabling seamless integration with existing VLA frameworks.
Please refer to \suppltext{\ref{app:sem_tok}} for the semantic training framework and the corresponding preprocessing illustrations.


\begin{figure}[tb]
    \centering
    \includegraphics[width=0.98\linewidth]{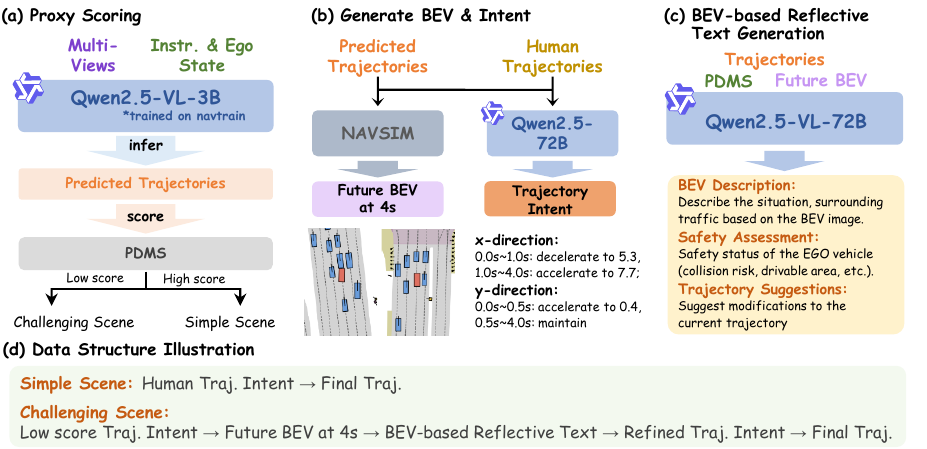}
    \caption{
   \textbf{Adaptive multimodal reflection data construction.} 
(a) A lightly fine-tuned planner is used to split the NAVSIM navtrain set into challenging and simple scenes based on predicted PDMS. 
(b) The simulator generates future BEV representations, while the LLM generates trajectory intents. 
(c) A VLM generates BEV-grounded reflective text. 
(d) The resulting data structures for simple and challenging scenes.
    }
    \label{fig:create_data}
\end{figure}

\subsection{Data Preparation}
\label{subsec-data}
To equip the VLM with driving scene understanding, BEV semantic forecasting, and adaptive multimodal reflection capabilities, we perform a data preparation stage as follows:
\begin{enumerate}[label=\textbf{\textit{\alph*)}}, leftmargin=*, nosep]
    \item \textbf{\textit{Pretraining Data:}}
    To adapt VLMs to autonomous driving scenarios, we leverage diverse multimodal datasets for both driving scene understanding and BEV modeling. 
    Specifically, we incorporate open-source understanding datasets, including DriveLM~\cite{drivelm}, NuInstruct~\cite{nuinstruct}, nuScenesQA~\cite{nusceneqa}, CODA-LM~\cite{coda-lm}, and others~\cite{lingoqa,drivegpt4,recogdrive}. 
    In addition, we construct BEV reconstruction and prediction data using the NAVSIM v1~\cite{navsimv1} simulator. 
    The reconstruction data maps current multi-view observations to BEV representations, capturing the present spatial layout. 
    The prediction data maps observations and future trajectories to future BEV representations, enabling trajectory-conditioned scene forecasting. 
    
    \item \textbf{\textit{Supervised Fine-Tuning Data:}}
To endow the model with trajectory planning capability, we utilize the official NAVSIM v1 navtrain split, denoted as $\mathcal{D}_{traj}$.
Each sample takes scene context and ego states as inputs and provides future trajectories over a 4-second horizon with a 0.5-second sampling interval as outputs.
To support adaptive multimodal reflection, we further construct reflection data from $\mathcal{D}_{traj}$.
As shown in Fig.~\ref{fig:create_data}, we first train a base VLA on $\mathcal{D}_{traj}$ and use its predicted PDMS scores to split the data into challenging and simple scenes.
The bottom 4,000 low-scoring samples are labeled as challenging scenes ($\mathcal{D}_{refl}^{+}$), while the remaining samples are treated as simple scenes ($\mathcal{D}_{refl}^{-}$).
For each sample, we use the predicted trajectory to generate an intent-conditioned future BEV with the NAVSIM v1 simulator, and use the human trajectory to generate the target trajectory intention with an LLM.
For low-scoring samples, failures such as collisions or insufficient safety margins become spatially observable in the predicted future BEV, providing grounded evidence for reflection.
We then feed the trajectory, future BEV representation, and corresponding PDMS score into an open-source VLM to generate the textual reflection component.
    The data structure of the final reflection dataset is shown in Fig.~\ref{fig:create_data} (d).
    For simple scenes, the model infers the intention and generates the trajectory. 
    For challenging scenes, the model infers the intention, performs multimodal reflection, refines it, and then generates the final trajectory.
    For detailed data examples, please refer to \suppltext{\ref{app:data_exp}}.

    \item \textbf{\textit{RL Training Data:}}
    We observe that during the RL training stage, a substantial portion of rollout sequences receive maximal rewards, resulting in a lack of discriminative training signals.
    To address this, we introduce a score-and-variance-based sampling strategy to select more informative training samples.
    Specifically, we estimate the reward distribution of each sample in $\mathcal{D}_{refl}$ using the SFT-trained model under stochastic rollouts. 
    Based on the resulting score statistics, we first select samples with low rewards, and then identify those with high score variance as informative instances to form the subset $\mathcal{D}_{refl}^{rl}$.
    The pseudocode is available in \suppltext{\ref{app:data}}.
    
\end{enumerate}

\begin{figure}[bt]
    \centering
    \includegraphics[width=0.98\linewidth]{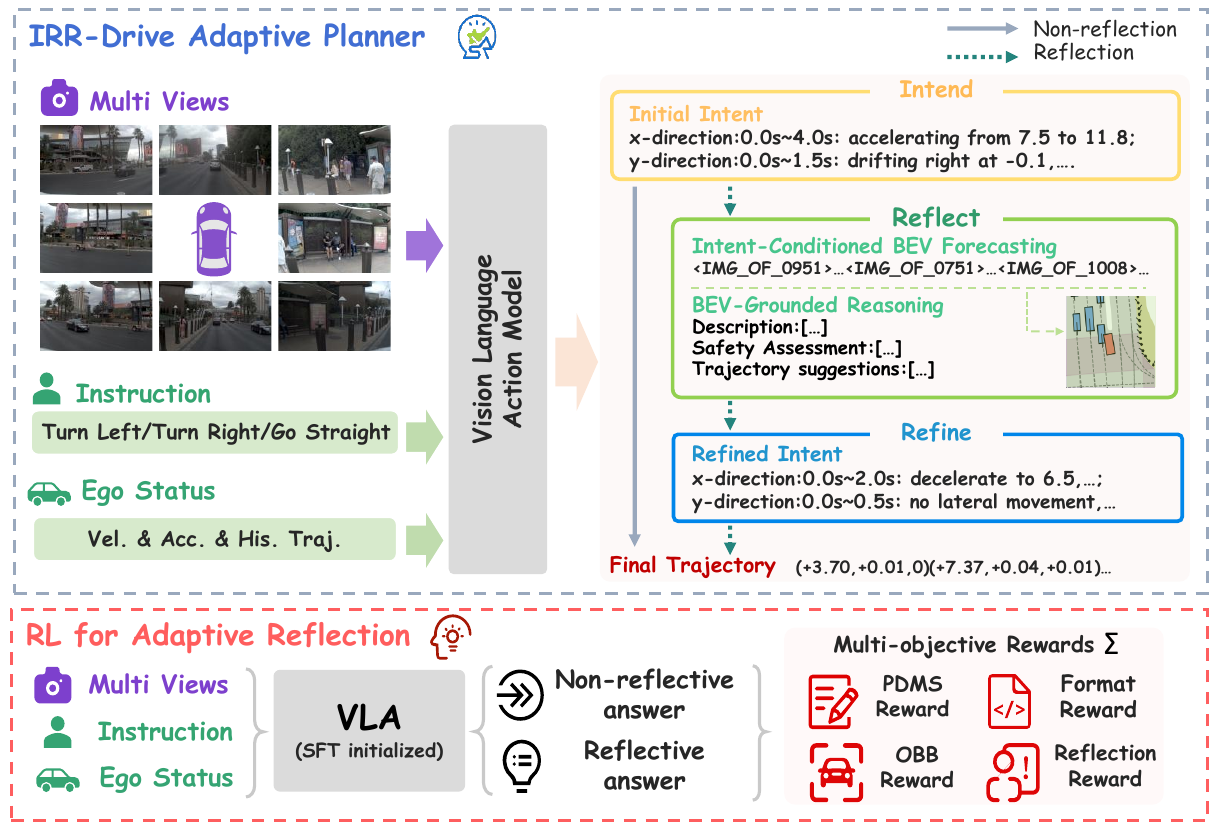}
    \caption{
    We present \textbf{\method{}}, an end-to-end autonomous driving framework that adaptively selects between “Non-Reflection” and “Reflection” modes depending on scene complexity. 
    Within its reflection framework, it integrates visual and textual reasoning to refine intentions and trajectories. 
    In the reinforcement learning stage, multiple rewards, including PDMS, Format, and OBB rewards, are combined with the proposed adaptive reflection reward.
    }
    \label{fig:framework}
\end{figure}

\subsection{Two-Stage Supervised Fine-tuning}
\label{subsec-sft}

In the first stage, we fine-tune the model using the pretraining data from Sec.~\ref{subsec-data}, including driving VQA datasets and BEV reconstruction and prediction data. 
This stage enables the model to perceive drivable regions, understand road conditions, and learn the mapping from multi-view images to BEV.

In the second stage, we perform supervised fine-tuning on $\mathcal{D}_{traj}$ and $\mathcal{D}_{refl}$ (defined in Sec.~\ref{subsec-data}) to equip the model with capabilities for trajectory planning and adaptive multimodal reflection.
For each scene query $q=\{q_{cam}, q_{ego}, q_{his} \}$, outputs are supervised by $o^{traj}$ from $\mathcal{D}_{traj}$ and $o^{refl}$ from $\mathcal{D}_{refl}$, with the objective of maximizing conditional likelihood: 
\begin{equation}
    \mathcal{L}_{\text{SFT}} = \mathbb{E}_{(q,o) \sim \{\mathcal{D}^{traj},\mathcal{D}^{refl}}\} \left[ -\log \pi_\theta(o \mid q) \right].
\end{equation}

This stage enables model to produce diverse response modes for subsequent reinforcement learning.

\subsection{Adaptive Reflection via Reinforcement Learning}
\label{subsec-rl}

\begin{figure}[b]
    \centering
    \includegraphics[width=0.98\linewidth]{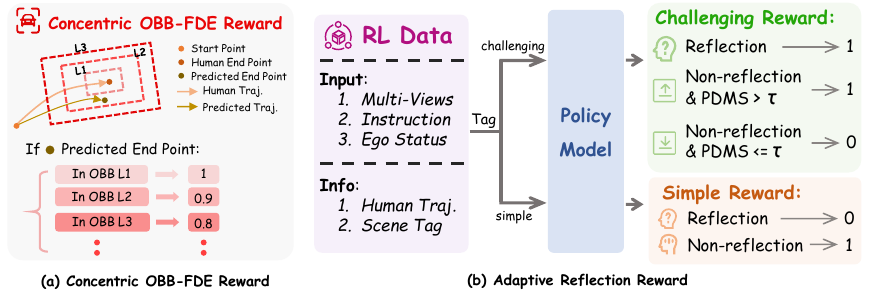}
    \caption{
    (a) The concentric OBB-FDE reward assigns tiered endpoint rewards using heading-aware bounding boxes around the expert endpoint. 
(b) The adaptive reflection reward encourages direct prediction in simple scenes, while allowing non-reflection in challenging scenes only when the current rollout already achieves sufficiently high PDMS.}
    \label{fig:reward}
\end{figure}

The two-stage supervised fine-tuning equips the model with an adaptive multimodal reflection mechanism, which directly drives its trajectory planning capabilities. 
However, imitation learning is constrained by predefined reflection labels and demonstration trajectories. 
Since these static targets may be suboptimal under closed-loop evaluation, the learned policy is limited to reproducing the training distribution and cannot explore alternative behaviors through interaction with the environment. 
To overcome this limitation and optimize the planner under closed-loop physical feedback, we further introduce reinforcement learning.

Due to the substantial length difference between reflective and non-reflective responses, token-level policy optimization can bias training toward longer reflective outputs and destabilize mode selection.
To mitigate this issue, we adopt GSPO~\cite{gspo}, which optimizes at the sequence level and provides a more balanced learning signal across reasoning modes.
Furthermore, our reward function is composed of four components:
\begin{enumerate}[label=\textbf{\textit{\alph*)}}, leftmargin=*]
    \item \textbf{\textit{PDMS Reward:}} 
    We directly utilize the PDMS score as the reward signal, which assesses trajectory safety and performance via the NAVSIM v1~\citep{navsimv1} simulator. 
    The score is represented as a continuous value ranging from 0 to 1, and is formulated as follows:
    \begin{equation}
        \text{PDMS} = \text{NC} \cdot \text{DAC} \cdot \frac{5 \cdot \text{TTC} + 5 \cdot \text{EP} + 2 \cdot \text{C}}{12}
        \label{eq:pdms}
    \end{equation}
    where PDMS integrates five sub-metrics: No At-Fault Collision (NC), Drivable Area Compliance (DAC), Time-to-Collision (TTC), Comfort (C), and Ego Progress (EP) to produce a comprehensive closed-loop planning score. 
    
    \item \textbf{\textit{Concentric OBB-FDE Reward:}}
    To evaluate spatial precision, we introduce a concentric OBB-FDE reward ($R_{obb}$), shown in Fig.~\ref{fig:reward} (a). 
    Instead of continuous distance penalties, this mechanism assigns a discrete, decaying reward based on the tightest bounding box from a set of concentrically scaled tiers (e.g., L1 for $1.0\times$ scale, L2 for $1.25\times$ scale) that encloses the predicted endpoint. 
    This stepwise design prevents overfitting to suboptimal human trajectories, enabling broader reinforcement learning exploration. 
    Crucially, by projecting the spatial deviation into the local coordinate frame of the ground truth pose to evaluate this required scaling factor against the ego-vehicle's asymmetric physical dimensions, specifically the front overhang ($l_{front}$), rear overhang ($l_{back}$), and half-width ($w_{half}$), $R_{obb}$ intrinsically accounts for anisotropic physical occupancy. 
    This provides a more accurate geometric constraint.
    Detailed computation of the Concentric OBB-FDE Reward is provided in the \suppltext{~\ref{app:obb_compute}}.

    \item \textbf{\textit{Adaptive Reflection Reward:}}
    To prevent the model from collapsing into a single mode of reasoning, we introduce an adaptive reflection reward ($R_{refl}$). 
    Since the scene tags in $D_{refl}$ are generated in an earlier stage, they are static and no longer aligned with the model’s current capabilities, making simple policy alignment unsuitable.
    In contrast, our design introduces a dynamic mechanism to adapt to changes in the model’s capabilities during training, as shown in Fig.~\ref{fig:reward} (b).
    Specifically, for simple scenes, we disable reflection and enforce a direct reasoning mode.
    For difficult scenes, however, we impose a score-based requirement: the model may use direct reasoning if the PDMS score exceeds $\tau$, and is otherwise required to perform reflection for refinement.
    Through this adaptive mechanism, the model learns to allocate its computational resources more intelligently, which improves both reasoning quality and inference efficiency.
    
    \item \textbf{\textit{Format Reward:}}
    The format reward $R_{format}$ enforces adherence to the predefined output format, including the correct placement and usage of all required tags as well as valid reflection and trajectory content.

\end{enumerate}

The overall reward function in the reinforcement learning stage is defined as follows:
\begin{equation}
    R = \lambda_1 (R_{\mathrm{PDMS}} + R_{\mathrm{OBB}}) 
+ \lambda_2 R_{\mathrm{refl}} 
+ \lambda_3 R_{\mathrm{format}}.
\end{equation}

where $  \lambda_1, \lambda_2, \lambda_3  $ denote the weight coefficients of each reward term, which balance their contributions to the total reward $  R  $.

Through these reward designs, the policy learns an adaptive reflection strategy during reinforcement learning and explores diverse alternatives around suboptimal human trajectories.
\section{Experiments}
\label{sec:exp}
\subsection{Experimental Setup}
\textbf{Dataset:} 
We perform comprehensive experiments on NAVSIM~\cite{navsimv1}, a data-driven non-reactive simulation benchmark for autonomous driving planning, built upon OpenScene.
In addition, the driving-domain pretraining data, the NAVSIM BEV reconstruction/prediction data, and the adaptive reflection data are constructed as described in Sec.~\ref{subsec-data}.

\textbf{Metric:} 
We evaluate our method on closed-loop benchmark NAVSIM v1~\cite{navsimv1} and NAVSIM v2~\cite{navsimv2}.
The NAVSIM v1 benchmark provides a nonreactive simulation environment and employs the Predictive Driver Model Score (PDMS) as its closed-loop planning metric, as defined in Equation~\ref{eq:pdms}.
Similarly, the NAVSIM v2 introduces the Extended Predictive Driver Model Score (EPDMS).
For the details of EPDMS, please refer to \suppltext{~\ref{app:epdms}}.


\textbf{Training Details:} 
We use Qwen2.5-VL-3B~\cite{qwen2.5vl} and Qwen3-VL-4B~\cite{qwen3vl} as the base model, with the corresponding semantic encoder used to initialize the encoder of the semantic BEV tokenizer.
The ablation experiments are conducted using Qwen3-VL-4B.
For more training details, please refer to \suppltext{~\ref{app:training_details}}.


\subsection{Performance Comparison}
\begin{table}[htbp]
    \centering
    \caption{
    Comparison on NAVSIM v1.
    Abbreviation: Diff. (Diffusion), Comf. (Comfort), Cam (Camera), L (LiDAR). 
    }
    \label{tab:navsimv1}
    
    \setlength{\tabcolsep}{4pt}
    \resizebox{0.98\textwidth}{!}{
    \begin{tabular}{lllcccccc}
        \toprule
        \textbf{Method} & \textbf{Backbone} & \textbf{Input} 
        & \textbf{NC$\uparrow$} & \textbf{DAC$\uparrow$} & \textbf{TTC$\uparrow$} 
        & \textbf{Comf.$\uparrow$} & \textbf{EP$\uparrow$} & \textbf{PDMS$\uparrow$} \\
        \midrule

        \multicolumn{9}{@{}l}{\raggedright \textit{Traditional End-to-End Methods}} \\
        
        VADv2~\cite{VADv2} & -- & Cam 
        & 97.2 & 89.1 & 91.6 & 100 & 76.0 & 80.9 \\

        TransFuser~\cite{TransFuser} & -- & Cam + L 
        & 97.7 & 92.8 & 92.8 & 100 & 79.2 & 84.0 \\

        Hydra-MDP++~\cite{Hydra-MDP++} & -- & Cam + L 
        & 98.3 & 96.0 & 94.6 & 100 & 78.7 & 86.5 \\

        Artemis~\cite{artemis} & -- & Cam 
        & 98.3 & 95.1 & 94.3 & 99.8 & 81.4 & 87.0 \\

        DiffusionDrive~\cite{DiffusionDrive} & -- & Cam + L 
        & 98.2 & 96.1 & 94.8 & 100 & 82.2 & 88.1 \\

        DriveDPO~\cite{drivedpo} & -- & Cam + L 
        & 98.5 & 98.1 & 94.8 & 99.9 & 84.3 & 90.0 \\

        \midrule

        \multicolumn{9}{@{}l}{\raggedright \textit{VLA Methods w/o Explicit Reasoning}} \\

        DrivingGPT~\cite{drivegpt4} & LLaMA2-7B~\cite{llama2} & Cam 
        & 98.1 & 90.7 & 94.9 & 95.6 & 79.7 & 82.4 \\

        Epona~\cite{Epona} & DiT-2.5B~\cite{dit25} & Cam 
        & 97.9 & 95.1 & 93.8 & 99.9 & 80.4 & 86.2 \\

        ReCogDrive~\cite{recogdrive} & InternVL3-8B~\cite{internvl3} & Cam 
        & 98.2 & 97.5 & 95.2 & 99.9 & 83.5 & 89.6 \\

        DriveVLA-W0~\cite{DriveVLA-W0} & Qwen2.5-VL-3B~\cite{qwen2.5vl} & Cam 
        & 98.7 & {99.1} & 95.3 & {99.3} & 83.3 & 90.3 \\

        \midrule

        \multicolumn{9}{@{}l}{\raggedright \textit{VLA Methods w/ Textual Reasoning}} \\

        AutoVLA~\cite{AutoVLA} & Qwen2.5-VL-3B~\cite{qwen2.5vl} & Cam 
        & 98.4 & 95.6 & {98.0} & 99.9 & 81.9 & 89.1 \\

        AdaThinkDrive~\cite{AdaThinkDrive} & InternVL3-8B~\cite{internvl3} & Cam 
        & 98.4 & 97.8 & 95.2 & 100 & 84.4 & 90.3 \\

        AutoDrive-$R^{2}$ ~\cite{autodriver2} & Qwen2.5-VL-7B~\cite{qwen2.5vl} & Cam 
        & 98.5 & 95.9 & 95.4 & 100 & 82.7 & 89.1 \\


        ELF-VLA-8B~\cite{elfvla} & InternVL3-8B~\cite{internvl3} & Cam 
        & {98.9} & 98.1 & {96.0} & 100 & 85.3 & {91.0} \\

        \midrule

        \multicolumn{9}{@{}l}{\raggedright \textit{VLA Methods w/ Visual Reasoning}} \\

        FSDrive~\cite{fsdrive} & Qwen2-VL-2B~\cite{qwen2vl} & Cam 
        & 98.2 & 93.8 & 93.3 & 99.9 & 80.1 & 85.1 \\

        DAP~\cite{dap} & MiMo-VL-7B~\cite{mimovl} & Cam+L
        & 95.2 & 97.4 & 96.5 & 100 & 82.2 & 87.2 \\

        PWM~\cite{pwm} & Show-o~\cite{showo} & Cam 
        & 98.6 & 95.9 & 95.4 & 100 & 81.8 & 88.1 \\
        
        LaST-VLA~\cite{last-vla} & InternVL3-8B~\cite{internvl3} & Cam 
        & {98.7} & 97.9 & 95.6 & 100 & {86.8} & {91.3} \\
        
        \midrule

        \colorrow{
        \textbf{\method{}-3B} & Qwen2.5-VL-3B~\cite{qwen2.5vl} & Cam 
        & 98.0 & 97.4 & 94.4 & 100 & 87.4 & 90.6 \\
        }
        \colorrow{
        \textbf{\method{}-4B} & Qwen3-VL-4B~\cite{qwen3vl} & Cam 
        & 98.0 & {98.3} & 93.7 & 100 & {88.5} & {91.3} \\
        }
        \bottomrule
    \end{tabular}
    }
\end{table}

\begin{table}[htbp]
    \centering
    \small
    \caption{
    Comparison with state-of-the-art methods on NAVSIM v2 with EPDMS.
    }
    \label{tab:navsimv2}
    
    \resizebox{0.98\textwidth}{!}{
    \begin{tabular}{lccccc|ccccc}
        \toprule
        \textbf{Method} 
        & \textbf{NC$\uparrow$} 
        & \textbf{DAC$\uparrow$} 
        & \textbf{DDC$\uparrow$} 
        & \textbf{TLC$\uparrow$} 
        & \textbf{EP$\uparrow$} 
        & \textbf{TTC$\uparrow$} 
        & \textbf{LK$\uparrow$} 
        & \textbf{HC$\uparrow$} 
        & \textbf{EC$\uparrow$} 
        & \textbf{EPDMS$\uparrow$} \\
        \midrule

        \multicolumn{11}{@{}l}{\textit{Traditional End-to-End Methods}} \\

        HydraMDP++ ~\cite{Hydra-MDP++}
        & 97.2 & 97.5 & 99.4 & 99.6 & 83.1 
        & 96.5 & 94.4 & 98.2 & 70.9 & 81.4 \\

        DriveSuprim ~\cite{drivesuprim} 
        & 97.5 & 96.5 & 99.4 & 99.6 & 88.4 
        & 96.6 & 95.5 & 98.3 & 77.0 & 83.1 \\

        DiffusionDrive ~\cite{DiffusionDrive}
        & 98.2 & 95.9 & 99.4 & 99.8 & 87.5 
        & 97.3 & 96.8 & 98.3 & 87.7 & 84.5 \\

        \midrule

        \multicolumn{11}{@{}l}{\textit{VLA Methods w/o Explicit Reasoning}} \\

        RecogDrive-8B ~\cite{recogdrive}
        & 98.3 & 95.2 & 99.5 & 99.8 & 87.1 
        & 97.5 & 96.6 & 98.3 & 86.5 & 83.6 \\

        WAM-Flow-1.5B ~\cite{wamflow}
        & 98.5 & 94.5 & 99.5 & 99.8 & 86.9 
        & 96.8 & {97.4} & 97.6 & 73.9 & 84.7 \\

        DriveVLA-W0-3B ~\cite{DriveVLA-W0}
        & 98.5 & {99.1} & 98.0 & 99.7 & 86.4 
        & 98.1 & 93.2 & 97.9 & 58.9 & 86.1 \\

        \midrule

        \multicolumn{11}{@{}l}{\textit{VLA Methods w/ Textual Reasoning}} \\

        ELF-VLA-8B ~\cite{elfvla}
        & {98.9} & 98.1 & 99.4 & 99.8 & 88.5 
        & {98.4} & 96.9 & 98.3 & 87.2 & {87.1} \\

        Senna-2-3B ~\cite{senna2}
        & 98.5 & 97.8 & 99.5 & 99.8 & 88.1 
        & 97.5 & {97.0} & {98.6} & 88.4 & 86.6 \\

        \midrule

        \multicolumn{10}{@{}l}{\raggedright \textit{VLA Methods w/ Visual Reasoning}} \\
        LaST-VLA-8B~\cite{last-vla}
        & {98.7} & 97.9 & 99.2 & 99.7 & {90.3} 
        & {98.2} & 96.6 & 98.3 & 86.3 & {87.1} \\
        \midrule

        \colorrow{
        \textbf{\method{}-3B}   
        & 97.3 & 97.4 & 98.7 & 99.6 & 91.9 
        & 97.1 & 95.7 & 97.9 & 76.9 & {87.9} \\
        }

        \colorrow{
        \textbf{\method{}-4B}  
        & 97.0 & {98.3} & 98.9 & 99.5 & {92.3} 
        & 96.8 & 95.8 & 97.6 & 82.2 & {89.0} \\
        }

        \bottomrule
    \end{tabular}
    }
\end{table}

\textbf{NAVSIM Benchmark.} 
Tab.~\ref{tab:navsimv1} and Tab.~\ref{tab:navsimv2} present comparisons of our method with others on the NAVSIM benchmark.
On NAVSIM v1, our method achieves competitive performance, reaching 91.3 PDMS, outperforming traditional end-to-end methods and recent VLA methods. 
In addition to PDMS, our model consistently improves trajectory-related metrics such as DAC and EP, indicating enhanced decision consistency and control quality.
On the more challenging NAVSIM v2 benchmark, our advantages become more pronounced. 
\method{} achieves 89.0 EPDMS, surpassing the previous best result (e.g., ELF-VLA-8B, 87.1) by a clear margin of +1.9. 
Meanwhile, we achieve the highest EP (92.3), demonstrating stronger task-completion capability in complex scenes.
These results suggest that our method achieves stronger generalization and robustness.

\begin{table}[htbp]
\centering
\caption{
Comparison with different reasoning modes.
\textbf{Runtime} denotes the average inference time per sample on navtest, measured on a single H800 GPU.
}
\setlength{\tabcolsep}{3pt}
\label{tab:mode}
    \resizebox{0.92\textwidth}{!}{
    \begin{tabular}{l | ccc ccc | c}
        \toprule
        \textbf{Model} & \textbf{NC$\uparrow$} & \textbf{DAC$\uparrow$} & \textbf{TTC$\uparrow$} & \textbf{Comf.$\uparrow$} & \textbf{EP$\uparrow$} & \textbf{PDMS$\uparrow$} & \textbf{Runtime (s)} \\
        \midrule
        Non-Reflection  
        & 98.1 & 97.9 & 93.4 & 100 & 87.6 & 90.6 & 1.46 \\
        
        Reflection 
        & 98.5 & 97.9 & 95.2 & 100 & 86.2 & 90.8 & 3.03 \\

        \midrule
        
        \graycolorrow{
        \textbf{\method{}} & 98.0 & 98.3 & 93.7 & 100 & 88.5 & 91.3 & 1.70 \\
        }
        \bottomrule
    \end{tabular}
    }
\end{table}
\textbf{Adaptive Reflection Performance.} 
We modify the format reward function in the reinforcement learning stage to enforce single-mode reasoning during exploration.
As shown in Tab.~\ref{tab:mode}, applying a uniform Reflection mode across all scenarios incurs a substantial computational overhead, increasing the runtime from 1.46s to 3.03s while yielding only a marginal +0.2 improvement in PDMS over the Non-Reflection baseline. 
This result highlights the inefficiency of indiscriminate reflective reasoning. 
In contrast, IRR-Drive adaptively allocates computation only when necessary.
Consequently, it achieves the best trade-off, reaching 91.3 PDMS with only 1.70s runtime, demonstrating both the effectiveness and efficiency of the proposed adaptive reflection reward function.

\subsection{Ablation Studies}

\begin{table}[htbp]
\centering
\small
\caption{
    \textbf{Ablation on Reward Functions.}
    \textbf{Refl. Ratio} denotes the proportion of samples that adopt reflection during inference on the navtest.
}
\label{tab:reward_ablation}
    \resizebox{0.98\textwidth}{!}{
        \begin{tabular}{c | c c c | ccc ccc | c}
            \toprule
            \textbf{ID} & \textbf{PDMS} & \makecell{\textbf{Adaptive} \\ \textbf{Reflection}} &  \makecell{\textbf{Concentric} \\ \textbf{OBB-FDE}}  & \textbf{NC$\uparrow$} & \textbf{DAC$\uparrow$} & \textbf{TTC$\uparrow$} & \textbf{Comf.$\uparrow$} & \textbf{EP$\uparrow$} & \textbf{PDMS$\uparrow$} & \makecell{\textbf{Refl.} \\ \textbf{Ratio}} \\
            \midrule
            
            1 & & & 
            & 98.6 & 95.6 & 95.4 & 100 & 81.1 & 87.6 & 11.1\% \\
            
            2 & \checkmark & & 
            & 98.0 & 95.9 & 91.7 & 99.9 & 84.1 & 88.9 & 1.9\% \\
            
            3 & \checkmark & \checkmark & 
            & 96.9 & 97.4 & 91.1 & 99.9 & 90.8 & 90.5 & 22.9\% \\
            
            \rowcolor{gray!20}
            4 & \checkmark & \checkmark & \checkmark 
            & 98.0 & 98.3 & 93.7 & 100  & 88.5 & 91.3 & 20.6\% \\
            
            \bottomrule
        \end{tabular}
    }
\end{table}
\begin{table}[htbp]
\centering
\scriptsize
\caption{
    \textbf{Ablation on CoT content.} Text+BEV reasoning achieves the best performance after both SFT and RL, showing the complementarity between textual reasoning and BEV reflection.
}
\setlength{\tabcolsep}{3pt}
\label{tab:cot_content}
    \resizebox{0.9\textwidth}{!}{
    \begin{tabular}{cc | c | cccccc}
        \toprule
        \multirow{2}{*}{\textbf{Text}} & \multirow{2}{*}{\textbf{BEV}} & \textbf{SFT} & \multicolumn{6}{c}{\textbf{RL}} \\[-3pt]
        \cmidrule(lr){3-3} \cmidrule(lr){4-9} 
         & & \textbf{PDMS$\uparrow$} & \textbf{NC$\uparrow$} & \textbf{DAC$\uparrow$} & \textbf{TTC$\uparrow$} & \textbf{Comf.$\uparrow$} & \textbf{EP$\uparrow$} & \textbf{PDMS$\uparrow$}  \\[-2pt]
         
        \midrule
         \ding{55} & \ding{55} 
         & 87.3 
         & 97.5 & 97.0 & 93.0 & 100 & 87.4 & 89.9 \\[-2pt]
         
         \ding{51} & \ding{55} 
         & 87.2 
         & 97.9 & 97.6 & 93.5 & 100 & 87.9 & 90.6 \\[-2pt]
         
         \ding{55} & \ding{51}
         & 87.2 
         & 98.8 & 97.5 & 96.4 & 100 & 84.8 & 90.5 \\[-2pt]
         
         \rowcolor{gray!20} 
         \ding{51} & \ding{51} 
         & 87.6 
         & 98.0 & 98.3 & 93.7 & 100 & 88.5 & 91.3 \\[-2pt]
        \bottomrule
    \end{tabular}
    }
\end{table}

\begin{table}[htbp]
    \centering
    \small
    \setlength{\tabcolsep}{2.5pt}
    
    \begin{minipage}[t]{0.46\textwidth}
        \centering
        \caption{
            \textbf{Ablation on Group Size.} A group size of 8 yields the best performance, indicating a favorable trade-off between rollout diversity and training efficiency.
        }
        \label{tab:groupsize}
        \resizebox{\linewidth}{!}{
            \begin{tabular}{@{} c | ccc ccc @{}}
                \toprule
                \textbf{Group Size} & \textbf{NC$\uparrow$} & \textbf{DAC$\uparrow$} & \textbf{TTC$\uparrow$} & \textbf{Comf.$\uparrow$} & \textbf{EP$\uparrow$} & \textbf{PDMS$\uparrow$} \\
                \midrule
                w/o RL  
                & 98.6 & 95.6 & 95.4 & 100 & 81.1 & 87.6 \\
                
                4 
                & 97.7 & 97.7 & 92.7 & 99.9 & 86.2 & 90.8 \\

                \graycolorrow{
                8 & 98.0 & 98.3 & 93.7 & 100 & 88.5 & 91.3 \\
                }

                12
                & 98.5 & 97.9 & 95.2 & 100 & 86.2 & 91.2 \\
                \bottomrule
                
            \end{tabular}
        }
    \end{minipage}\hfill
    \begin{minipage}[t]{0.52\textwidth}
        \centering
        \caption{
        \textbf{Ablation on Training Stages.} Pretraining improves generalization over SFT alone, while RL fine-tuning further enhances closed-loop planning performance, leading to the best overall result.
        }
        \label{tab:training_strategy}
        \resizebox{\linewidth}{!}{
            \begin{tabular}{@{} l | ccc ccc @{}}
                \toprule
                \textbf{Model} & \textbf{NC$\uparrow$} & \textbf{DAC$\uparrow$} & \textbf{TTC$\uparrow$} & \textbf{Comf.$\uparrow$} & \textbf{EP$\uparrow$} & \textbf{PDMS$\uparrow$} \\
                \midrule
                SFT 
                & 97.7 & 93.3 & 93.9 & 100 & 80.5 & 85.7 \\
                
                Pre+SFT 
                & 98.6 & 95.6 & 95.4 & 100 & 81.1 & 87.6 \\
        
                \graycolorrow{
                Pre+SFT+RL 
                & 98.0 & 98.3 & 93.7 & 100 & 88.5 & 91.3 \\
                }
                \bottomrule
            \end{tabular}
        }
    \end{minipage}
    
\end{table}

\textbf{Ablation on Reward Functions.}
Tab.~\ref{tab:reward_ablation} illustrates the impact of different reward combinations on closed-loop metrics. 
Experiment 1 reflects the pre-RL model performance and is adopted as the baseline.
With only the basic PDMS reward, the model achieves a PDMS score of 88.9, representing an improvement of only 1.3 over the baseline.
Additionally, its reasoning mode collapsed, with a reflection rate of only 1.9\%.
Adding the adaptive reflection reward significantly improves the score to 90.5, and its adaptive reflection mode is effectively explored, achieving a 22.9\% reflection rate. 
Incorporating our proposed Concentric OBB-FDE reward further elevates the PDMS metric to a peak of 91.3, while maintaining a stable reflection rate of 20.6\%.
These results demonstrate that the Concentric OBB-FDE reward effectively strikes an optimal balance between accommodating sub-optimal human trajectories and facilitating RL exploration, thereby further enhancing the model's performance.

\textbf{Ablation on CoT Content.}
Tab.~\ref{tab:cot_content} presents an ablation study on the impact of different CoT content on NAVSIM v1. 
Results show that relying on a single modality (either Text or BEV) yields limited gains in the RL stage.
In contrast, combining Text and BEV yields the best overall performance. 
The multimodal CoT achieves the highest PDMS in the SFT stage (87.6) and further benefits from RL fine-tuning, reaching 91.3. 
These results suggest that the spatial geometric information from BEV and the reasoning capability of text are complementary, and their integration is crucial for improving CoT effectiveness in trajectory planning.

\textbf{Ablation on Group Size.}
As shown in Tab.~\ref{tab:groupsize}, while applying RL significantly elevates the PDMS over the baseline (87.6), the performance gain saturates as the group size increases. Specifically, the PDMS improves from 90.8 to 91.3 when scaling the group size from 4 to 8, but slightly drops to 91.2 at a group size of 12. 
This demonstrates that a group size of 8 is sufficient to achieve optimal planning capability without redundant computational overhead.

\textbf{Ablation on Training Stage.}
Tab.~\ref{tab:training_strategy} validates the effectiveness of the proposed multi-stage training pipeline. 
Incorporating pretraining into the SFT baseline consistently improves generalization (PDMS +1.9).
Further introducing reinforcement learning leads to substantial improvements in decision-making performance, with notable gains in DAC and EP. 
The full pipeline (Pre+SFT+RL) achieves the best overall performance, reaching a PDMS of 91.3 (+5.6 over SFT), thereby highlighting the critical role of RL in closed-loop execution.


More ablation studies on \textit{High Score Thresholds($\tau$)} and \textit{Training Data in RL} are provided in \suppltext{~\ref{app:abl_hst_data}}.


\section{Conclusion}
\label{sec:ccl}
In this paper, we argue that existing reasoning paradigms for autonomous driving are weakly grounded and treat reflection as a static, post-hoc process. 
To address this, we propose IRR-Drive, a novel multimodal reflection framework that leverages predicted future semantic BEV representations as a structured space for plan verification. 
In addition, we enable adaptive reflection, allowing the model to dynamically adjust reasoning based on scene complexity through a reinforcement learning pipeline.
Experiments on NAVSIM v1 and v2 demonstrate the effectiveness of IRR-Drive, showing that multimodal reasoning is complementary and that adaptive reflection balances inference efficiency and performance.

\textbf{Limitation and Future Work.}
Although our model achieves competitive performance and incorporates an adaptive reasoning paradigm, it is still not fully suitable for real-time inference. 
Future work will focus on real-time deployment and improving inference efficiency (e.g., through model quantization and distillation).

{
\small
\bibliographystyle{plainnat}
\bibliography{neurips_2026}
}


\newpage
\appendix
\section{Supplementary Material}

\subsection{More Related Work}
\label{app:related_work}

\textbf{Vision-Language and Vision-Language-Action Models for Autonomous Driving.} 
The integration of large vision-language models into autonomous driving enables semantic-level decision-making grounded in scene understanding and commonsense knowledge. 
DriveGPT4 ~\cite{drivegpt4} is among the earliest efforts to jointly model driving interpretation and action prediction within a unified vision-language framework. 
To address the limited precision of language models in continuous planning, Senna ~\cite{senna} decouples semantic decision-making from trajectory prediction, using language-based meta-actions as the interface between a high-level LVLM and a low-level planner. 
ReCogDrive ~\cite{recogdrive} further addresses domain gap, language--action mismatch, and imitation bias through a three-stage pipeline consisting of driving VQA pretraining, a cognitive-guided diffusion planner, and reinforcement learning fine-tuning. 
These works demonstrate the promise of the VLA paradigm for autonomous driving, while most of them still rely on single-pass generation without an explicit feedback mechanism for validating the physical consequences of planned behaviors.

\textbf{Future Scene Modeling for Autonomous Driving.} 
Another closely related line of work incorporates future prediction or world modeling into driving planning. 
ReasonPlan ~\cite{ReasonPlan} combines self-supervised next-scene prediction with decision chain-of-thought to jointly model scene forecasting and decision reasoning in closed-loop driving. 
Epona ~\cite{Epona} proposes an autoregressive diffusion world model for long-horizon driving video generation and applies the learned world model to planning. 
DriveVLA-W0 ~\cite{DriveVLA-W0} addresses the supervision deficit in VLA training by using future image prediction as a dense self-supervised signal. 
DriveWorld-VLA ~\cite{DriveWorld-VLA} further couples world dynamics with the planner in latent space, enabling decision-making to directly leverage latent representations of future scene evolution. 
Collectively, these methods incorporate future scenario modeling in different forms, including auxiliary supervision, reasoning support, and latent planning.

\subsection{Semantic Tokenizer}
\label{app:sem_tok}
\begin{figure}[htbp]
    \centering
    \includegraphics[width=0.95\linewidth]{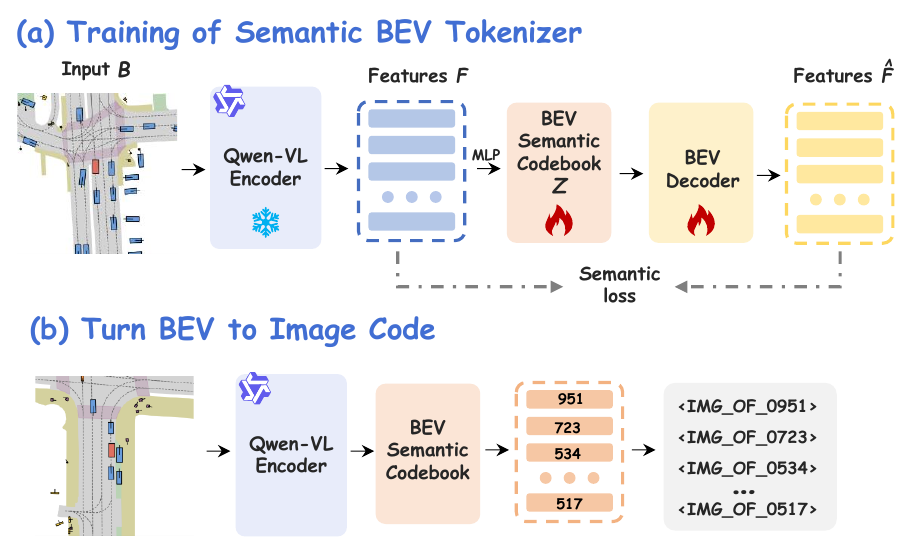}
    \caption{
    (a) Training framework of the semantic BEV tokenizer. 
    A frozen semantic encoder processes the BEV representation to extract features, which are then discretized and decoded to reconstruct the continuous semantic features.
    (b) The trained semantic BEV tokenizer is used to convert BEV representations into discrete codes. To align with the autoregressive workflow of the VLM, these codes are further mapped into special tokens in the format <IMG\_OF\_XXXX>.
    }
    \label{fig:sem_tok}
\end{figure}

As shown in Fig.~\ref{fig:sem_tok}, we demonstrate how to train a semantic BEV tokenizer and how to use it to convert BEV maps into discrete special tokens compatible with the autoregressive VLM interface.

\subsection{Adaptive Data Examples}

\begin{figure}[htbp]
    \centering
    \begin{minipage}{\linewidth}
        \centering
        \includegraphics[width=0.98\linewidth]{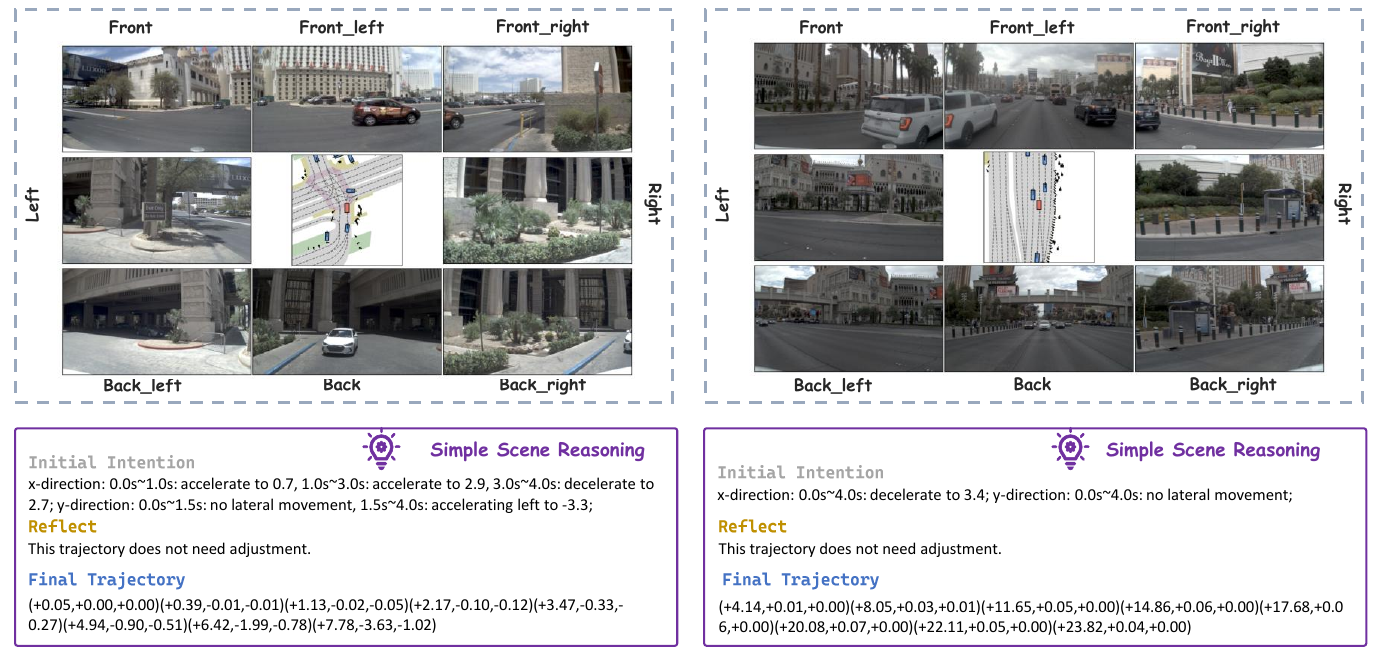}
        \caption{Illustrative examples of simple scenes.}
        \label{fig:data_simple}
    \end{minipage}
    
    \vspace{1.5em} 
    
    \begin{minipage}{\linewidth}
        \centering
        \includegraphics[width=0.98\linewidth]{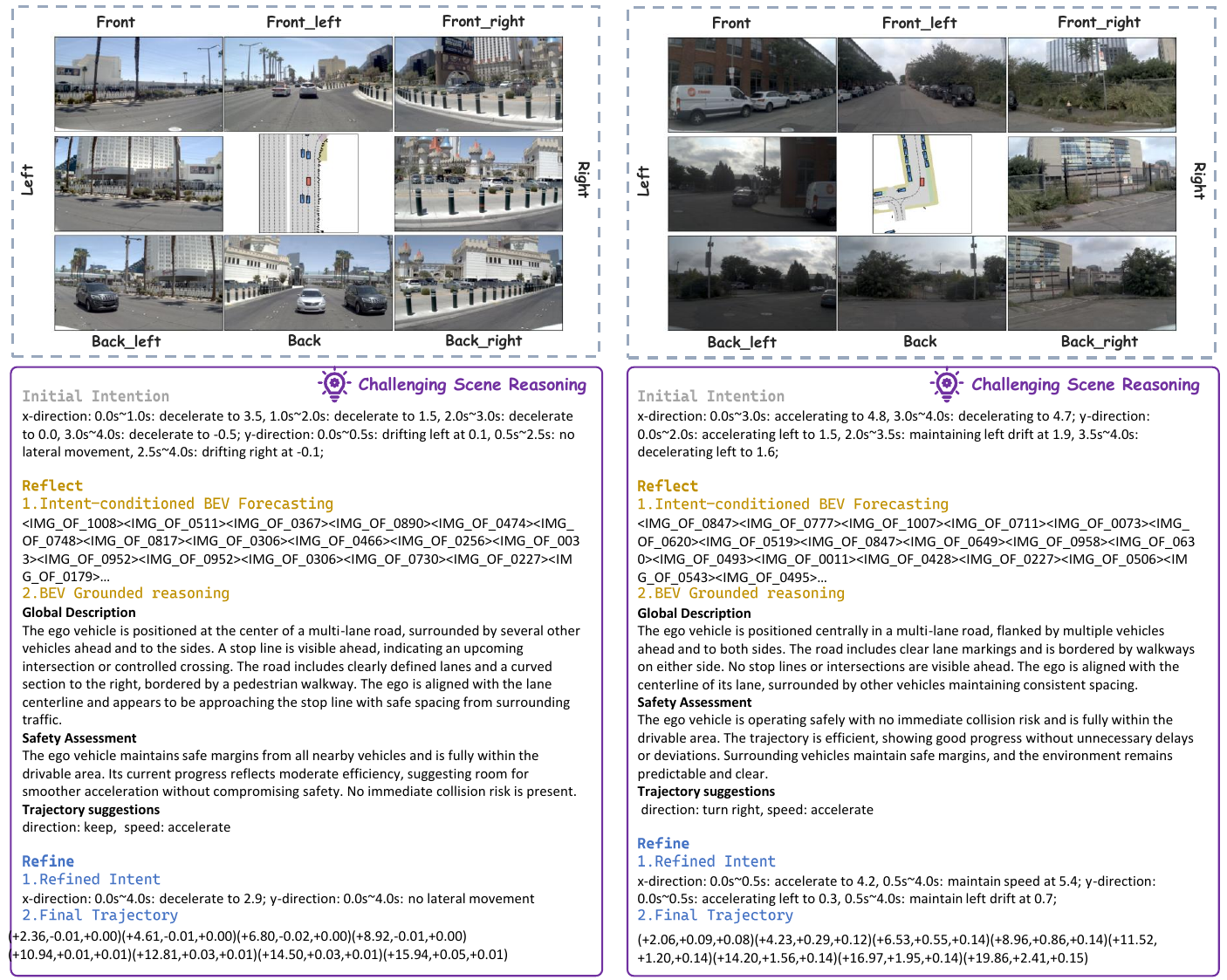}
        \caption{Illustrative examples of challenging scenes.}
        \label{fig:data_challenging}
    \end{minipage}
\end{figure}

We present examples of data from simple and challenging scenarios in Fig.~\ref{fig:data_simple} and ~\ref{fig:data_challenging}, respectively.

\label{app:data_exp}

\subsection{Pseudocode for data construction and filtering}
\label{app:data}
We present the pseudocode for data construction, as shown in Algs~\ref{alg:data_preparation} and ~\ref{alg:rl_filter}.

\begin{algorithm}[b]
\caption{Data Preparation for Reflection-Augmented Training}
\label{alg:data_preparation}
\small
\begin{algorithmic}[1]
\State \textbf{Input:} driving understanding data $\mathcal{D}_{under}$, NAVSIM, trajectory data $\mathcal{D}_{traj}$, base planner $\mathcal{M}$, VLM $\mathcal{V}$, LLM $\mathcal{L}$, rollout number $K=8$, temperature $\tau_{\mathrm{temp}}=1$, hard-sample number $N_{hard}=4000$
\State \textbf{Output:} pretraining data $\mathcal{D}_{pre}$, reflection data $\mathcal{D}_{refl}$, RL data $\mathcal{D}_{refl}^{rl}$

\State Construct $\mathcal{D}_{pre}$ from $\mathcal{D}_{under}$, BEV reconstruction data, and BEV prediction data
\State Train $\mathcal{M}$ on $\mathcal{D}_{traj}$
\For{each sample $x \in \mathcal{D}_{traj}$}
    \State Predict trajectory $\hat{\mathbf{y}} \leftarrow \mathcal{M}(x)$
    \State Compute PDMS score $s$ with NAVSIM
\EndFor
\State Rank all samples by PDMS score
\State Define the bottom $N_{hard}$ samples as $\mathcal{D}_{refl}^{+}$ and the remaining ones as $\mathcal{D}_{refl}^{-}$

\For{each sample $x \in \mathcal{D}_{traj}$}
    \State Obtain ground-truth trajectory $\mathbf{y}^{*}$
    \State Generate trajectory intention $\mathbf{z} \leftarrow \mathcal{L}(\mathbf{y}^{*})$
    \If{$x \in \mathcal{D}_{refl}^{+}$}
        \State Generate future BEV $\mathbf{B}^{f}$ with NAVSIM conditioned on $\hat{\mathbf{y}}$
        \State Generate textual reflection $r \leftarrow \mathcal{V}(\hat{\mathbf{y}}, \mathbf{B}^{f}, s)$
        \State Add reflective sample $(x, \mathbf{z}, \mathbf{B}^{f}, r, \mathbf{y}^{*})$ to $\mathcal{D}_{refl}$
    \Else
        \State Add non-reflective sample $(x, \mathbf{z}, \mathbf{y}^{*})$ to $\mathcal{D}_{refl}$
    \EndIf
\EndFor

\State Train an SFT model on $\mathcal{D}_{traj} \cup \mathcal{D}_{refl}$
\For{each sample $d \in \mathcal{D}_{refl}$}
    \State Perform $K$ rollouts with temperature $\tau_{\mathrm{temp}}$
    \State Compute rollout PDMS scores
    \State Compute mean score $\mu_d$ and variance $\sigma_d^2$
\EndFor
\State Select samples with low mean score and high variance to form $\mathcal{D}_{refl}^{rl}$

\State \textbf{return} $\mathcal{D}_{pre}, \mathcal{D}_{refl}, \mathcal{D}_{refl}^{rl}$
\end{algorithmic}
\end{algorithm}

\begin{algorithm}[htbp]
\caption{Informative Sample Selection for RL Training}
\label{alg:rl_filter}
\small
\begin{algorithmic}[1]
\State \textbf{Input:} Reflection dataset $\mathcal{D}_{refl}$, SFT model $\mathcal{M}_{sft}$, rollout number $K=8$, temperature $\tau=1$
\State \textbf{Output:} RL training subset $\mathcal{D}_{refl}^{rl}$

\State Initialize $\mathcal{D}_{refl}^{rl} \gets \emptyset$

\For{each sample $d \in \mathcal{D}_{refl}$}
    \State Sample $K$ rollout trajectories $\{\hat{\mathbf{y}}_1, \ldots, \hat{\mathbf{y}}_K\}$ from $\mathcal{M}_{sft}$ with temperature $\tau$
    \State Evaluate PDMS scores $\{s_1, \ldots, s_K\}$
    \State Compute mean score $\mu_d \gets \frac{1}{K}\sum_{k=1}^{K} s_k$
    \State Compute variance $\sigma_d^2 \gets \frac{1}{K}\sum_{k=1}^{K}(s_k-\mu_d)^2$
\EndFor

\State Select samples with low $\mu_d$ and high $\sigma_d^2$
\State Add selected samples into $\mathcal{D}_{refl}^{rl}$

\State \textbf{return} $\mathcal{D}_{refl}^{rl}$
\end{algorithmic}
\end{algorithm}

\subsection{Detailed computation of the Concentric OBB-FDE Reward}
\label{app:obb_compute}
We provides the formal mathematical definition of the concentric OBB-FDE reward ($R_{obb}$) introduced in the main text. The calculation is decoupled into two stages: computing the geometric spatial deviation and mapping it to a discrete reward score.

\textbf{Spatial Deviation Calculation}
Let $\mathbf{P}_{gt} = (x_{gt}, y_{gt}, \theta_{gt})$ denote the ground truth pose of the ego-vehicle at the final trajectory timestep, and $\mathbf{P}_{res} = (x_{res}, y_{res})$ denote the predicted endpoint. We first calculate the global positional error and project it into the local coordinate frame of the ground truth pose:
\begin{equation}
\begin{bmatrix} x_{local} \\ y_{local} \end{bmatrix} = 
\begin{bmatrix} \cos \theta_{gt} & \sin \theta_{gt} \\ -\sin \theta_{gt} & \cos \theta_{gt} \end{bmatrix}
\begin{bmatrix} x_{res} - x_{gt} \\ y_{res} - y_{gt} \end{bmatrix}.
\end{equation}

To account for the asymmetric physical dimensions of the ego-vehicle, we define the front overhang as $l_{front}$, the rear overhang as $l_{back}$, and the half-width as $w_{half}$. The spatial deviation is quantified by the minimum concentric scaling factor $k$ required for the ego-vehicle's oriented bounding box to fully enclose the predicted endpoint. 

Instead of employing a piecewise function, this factor can be elegantly computed as the maximum required expansion ratio across both the longitudinal and lateral axes in a single unified expression:
\begin{equation}
k = \max \left( \frac{x_{local}}{l_{front}}, \frac{-x_{local}}{l_{back}}, \frac{|y_{local}|}{w_{half}} \right).
\end{equation}
This formulation naturally handles the longitudinal asymmetry, as the $\max()$ operator inherently filters out the irrelevant negative term depending on whether the prediction falls in front of or behind the vehicle's geometric center.

\textbf{Stepwise Reward Calculation}
Once the spatial scaling factor $k$ is obtained, it is mapped to a discrete reward score $R_{obb} \in [0, 1]$. We employ a stepwise decaying function based on predefined scale thresholds, representing concentrically expanded bounding boxes. The discrete reward is assigned as follows:
\begin{equation}
R_{obb}(k) = 
\begin{cases} 
1.0, & \text{if } k \le 1.0 \quad \text{(Within physical ego box)}\\ 
0.9, & \text{if } 1.0 < k \le 1.25 \\ 
0.8, & \text{if } 1.25 < k \le 1.5 \\ 
0.6, & \text{if } 1.5 < k \le 2.0 \\ 
0.4, & \text{if } 2.0 < k \le 3.0 \\ 
0.2, & \text{if } 3.0 < k \le 5.0 \\ 
0.0, & \text{otherwise (Severe deviation)} 
\end{cases}.
\end{equation}
This step-down mechanism avoids zero-sum binary penalties and provides a smooth, progressive learning signal during the reinforcement learning process while strictly adhering to the vehicle's physical collision boundaries.

\subsection{Detail of EPDMS}
\label{app:epdms}
The EPDMS metric is calculated as follows:
\begin{equation}
\begin{split}
    \mathrm{EPDMS}
=
\mathrm{NC}\cdot \mathrm{DAC}\cdot \mathrm{DDC}\cdot \mathrm{TLC}
\cdot
\left(
\frac{
5\mathrm{EP}+2\mathrm{LK}+2\mathrm{HC}+5\mathrm{TTC}+2\mathrm{EC}
}{16}
\right).
\end{split}
\end{equation}
where NC, DAC, Driving Direction Compliance (DDC), Traffic Light Compliance (TLC), EP, TTC, Lane Keeping (LK), History Comfort (HC), and Extended Comfort (EC) jointly provide a comprehensive evaluation of safety, comfort, and driving performance.

\subsection{More Training Details}
\label{app:training_details}
The model is trained in three stages, as described in Sec.~\ref{subsec-sft} and~\ref{subsec-rl}. 
In the first stage, we conduct SFT on pre-training data for 5 epochs with a learning rate of $2 \times 10^{-5}$ and batch size 128.
The second stage fine-tunes the model on supervised trajectory and reflection data (\(D_{\text{traj}}\), \(D_{\text{refl}}\)) for 5 epochs using a learning rate of $1\times 10^{-5}$, batch size 128.
In the third stage, we apply reinforcement learning for 3 epochs on $\mathcal{D}_{refl}^{rl}$, with a learning rate of $2\times 10^{-6}$, using a rollout size of 8 and a batch size of 128.
In addition, in the Concentric OBB-FDE Reward, the front overhang ($l_{front}$), rear overhang ($l_{back}$), and half-width ($w_{half}$) are set to 2.0, 1.2, and 1.0, respectively.
The default high score threshold $\tau$ for the Adaptive Reflection Reward is 0.92.
All experiments are conducted on the H800.

\subsection{Ablation on High Score Thresholds and Training Data in RL.}
\label{app:abl_hst_data}

\begin{table}[htbp]
    \centering
    \small
    \setlength{\tabcolsep}{2.5pt}
    \begin{minipage}[t]{0.50\textwidth}
        \centering
        \caption{
        \textbf{Ablation on high score thresholds.}
        \textbf{HST} means high score thresholds $\tau$.
        }
        \label{tab:rl_hs}
        \resizebox{\linewidth}{!}{
            \begin{tabular}{l | cccccc | c}
                \toprule
                \textbf{HST} & \textbf{NC$\uparrow$} & \textbf{DAC$\uparrow$} & \textbf{TTC$\uparrow$} & \textbf{Comf.$\uparrow$} & \textbf{EP$\uparrow$} & \textbf{PDMS$\uparrow$} & \makecell{\textbf{Refl.} \\ \textbf{Ratio}} \\
                \midrule
                0.90 & 98.0 & 98.2 & 93.1 & 100 & 87.9 & 90.2 & 14.9\%  \\
                \graycolorrow{
                0.92 & 98.0 & 98.3 & 93.7 & 100 & 88.5 & 91.3 & 20.6\% \\
                }
                0.94 & 98.2 & 98.0 & 94.2 & 100 & 88.1 & 91.2 & 29.1\% \\
                0.96 & 98.0 & 98.1 & 93.4 & 100 & 88.3 & 90.1 & 34.8\% \\
                \bottomrule
            \end{tabular}
        }
    \end{minipage}\hfill
    \begin{minipage}[t]{0.48\textwidth}
        \centering
        \caption{
            \textbf{Ablation study on the number of training data in RL.}
            \textsuperscript{*}: Sampled based on score and variance.
            \textsuperscript{$\dagger$}: Random Sampled.
        }
        \label{tab:rl_data}
        \resizebox{\linewidth}{!}{
            \begin{tabular}{l | cccccc }
                \toprule
                \textbf{Num.} & \textbf{NC$\uparrow$} & \textbf{DAC$\uparrow$} & \textbf{TTC$\uparrow$} & \textbf{Comf.$\uparrow$} & \textbf{EP$\uparrow$} & \textbf{PDMS$\uparrow$} \\
                \midrule
                132k & 96.2 & 97.3 & 93.4 & 100 & 88.5 & 87.7 \\
                15k\textsuperscript{$\dagger$} & 96.1 & 96.6 & 89.8 & 100 & 86.2 & 86.2 \\
                \graycolorrow{
                15k\textsuperscript{*}         & 98.0 & 98.3 & 93.7 & 100 & 88.5 & 91.3 \\
                }
                \bottomrule
            \end{tabular}
        }
    \end{minipage}
    
\end{table}
Tab.~\ref{tab:rl_hs} ablates the high score threshold ($\tau$). 
The results indicate that $\tau=0.92$ is the optimal operating point, yielding the highest PDMS of 91.3 while avoiding the computational overhead and performance degradation (90.1 PDMS) associated with excessive reflection ($\tau=0.96$).
Tab.~\ref{tab:rl_data} validates our score-and-variance-based sampling strategy during the RL stage. 
Fine-tuning with only 15k targeted informative samples (15k*) drastically outperforms not only the 15k random subset (86.2 PDMS) but also the full 132k dataset (87.7 PDMS), reaching 91.3 PDMS. 
This highlights that data quality and targeted exploration are far more crucial than sheer data volume for refining driving policies.

\subsection{Broader impacts}
\label{app:broader_impacts}
Autonomous driving is inherently safety-critical, a concern that becomes even more significant when language guidance is incorporated into VLA models. Such integration requires strong defenses against adversarial attacks, as well as mechanisms to detect and filter unsafe or malicious human instructions in advance. To reduce potential security risks, secure in-vehicle communication channels should be established, and model deployment should adopt a gated-release update strategy instead of relying on online continual reinforcement fine-tuning directly on individual vehicles.

\subsection{LLM usage}
\label{app:llm_usage}
As described in Sec~\ref{subsec-data}, we use open-source LLMs and VLMs to construct the dataset. 
In addition, we utilize closed-source models (GPT, Gemini) to assist in paper writing.


\end{document}